\documentclass[10pt,twocolumn,letterpaper]{article}

\usepackage{iccv}
\usepackage{times}
\usepackage{epsfig}
\usepackage{graphicx, overpic}
\usepackage{wrapfig,lipsum,booktabs}

\usepackage{enumitem}
\usepackage{xcolor}
\usepackage[breaklinks=true,colorlinks,citecolor=blue,urlcolor=blue,bookmarks]{hyperref}
\usepackage{amsmath}
\usepackage{amssymb}

\newcommand{\argmax}{\mathop{\mathrm{argmax}}\limits}

\newcommand{\ConfInf}{\vspace{-.7in} {\normalsize \normalfont \color{blue}{
   IEEE International Conference on Computer Vision (ICCV) 2019}} \vspace{.45in} \\}

\iccvfinalcopy 

\ifdefined \GramaCheck
  \newcommand{\CheckRmv}[1]{}
  \newcommand{\figref}[1]{Figure 1}%
  \newcommand{\tabref}[1]{Table 1}%
  \newcommand{\secref}[1]{Section 1}
  \newcommand{\equref}[1]{Equation 1}
\else
  \newcommand{\CheckRmv}[1]{#1}
  \newcommand{\figref}[1]{Fig.~\ref{#1}}%
  \newcommand{\tabref}[1]{Tab.~\ref{#1}}%
  \newcommand{\secref}[1]{Sec.~\ref{#1}}
  \newcommand{\equref}[1]{Eq.~(\ref{#1})}
\fi

\def\iccvPaperID{1406} 

\ificcvfinal\fi
\begin{document}

\title{\ConfInf Optimizing the F-measure for Threshold-free Salient Object Detection}

\author{
  Kai Zhao\textsuperscript{1}, Shanghua Gao\textsuperscript{1},
  Wenguan Wang\textsuperscript{2},
  Ming-Ming Cheng\textsuperscript{1}\thanks{M.M. Cheng is the corresponding author.}\\
  \textsuperscript{1}TKLNDST, CS, Nankai University~~~~ 
  \textsuperscript{2}Inception Institute of Artificial Intelligence\\
  {\tt\small \{kaiz.xyz,shanghuagao,wenguanwang.ai\}@gmail.com,cmm@nankai.edu.cn}
}

\maketitle
\thispagestyle{empty}

\begin{abstract}
  Current CNN-based solutions to salient object detection (SOD)
  mainly rely on the optimization of cross-entropy loss (CELoss).
  Then the quality of detected saliency maps is often evaluated in
  terms of F-measure.
  In this paper, we investigate an interesting issue:
  can we consistently use the F-measure formulation in both training and evaluation
  for SOD?
  By reformulating the standard F-measure, 
  we propose the \emph{relaxed F-measure} which is differentiable w.r.t 
  the posterior and can be easily appended to the back of CNNs as the loss function.
  Compared to the conventional cross-entropy loss of which the gradients decrease
  dramatically in the saturated area,
  our loss function, named FLoss, holds considerable gradients even when the activation
  approaches the target.
  Consequently, the FLoss can continuously force the network
  to produce polarized activations.
  Comprehensive benchmarks on several popular datasets show that FLoss
  outperforms the state-of-the-art with a considerable margin.
  More specifically, due to the polarized predictions,
  our method is able to obtain high-quality saliency maps without carefully tuning
  the optimal threshold, showing significant advantages in real-world applications.
  Code and pretrained models are available at \url{http://kaizhao.net/fmeasure}.
\end{abstract}

\section{Introduction}
We consider the task of salient object detection (SOD), where each pixel of a
given image has to be classified as salient (outstanding) or not.
The human visual system is able to perceive and process visual signals
distinctively: interested regions are conceived and analyzed with high priority
while other regions draw less attention.
This capacity has been long studied in the computer vision community
in the name of `salient object detection',
since it can ease the procedure of scene understanding~\cite{borji2015salient}.
The performance of modern salient object detection methods is often evaluated 
in terms of F-measure.
Rooted from information retrieval~\cite{van1974foundation},
the F-measure is widely used as an evaluation metric
in tasks where elements of a specified class have to be retrieved,
especially when the relevant class is rare.
Given the per-pixel prediction $\hat{Y} (\hat{y}_i \!\in\! [0, 1], i\!=\!1,...,|Y|)$
and the ground-truth saliency map $Y (y_i \!\in\! \{0, 1\}, i\!=\!1,...,|Y|)$,
a threshold $t$ is applied to obtain the binarized prediction
$\dot{Y}^t (\dot{y}^t_i \!\in\! \{0, 1\}, i\!=\!1,...,|Y|)$.
The F-measure is then defined as the harmonic mean of precision and recall:
\begin{equation}\small
\!\!F(Y, \dot{Y}^t) \!=\!
(1\!+\!\beta^2)\frac{\text{precision}(Y, \dot{Y}^t) \cdot \text{recall}(Y, \dot{Y}^t)}
{\beta^2 \text{precision}(Y, \dot{Y}^t) \!+\! \text{recall}(Y, \dot{Y}^t)},
\label{eq:def-f}
\end{equation}
where $\beta^2\!>\!0$ is a balance factor between precision and recall.
When $\beta^2\!>\!1$, the F-measure is biased in favour of recall
and otherwise in favour of precision.

Most CNN-based solutions for SOD
\cite{hou2017deeply,li2016deep,wang2016saliency,fan2019shifting,wang2019iterative,
zhao2019EGNet,wang2019salient}
mainly rely on the optimization of
\emph{cross-entropy loss} (CELoss) in an FCN~\cite{long2015fully} architecture,
and the quality of saliency maps is often assessed by the F-measure.
Optimizing the pixel-independent CELoss can be regarded as minimizing the mean absolute
error (MAE=$\frac{1}{N}\sum_i^N |\hat{y}_i - y_i|$), because in both circumstances
each prediction/ground-truth pair works independently and contributes
to the final score equally.
If the data labels have biased distribution, models trained with CELoss
would make biased predictions towards the majority class.
Therefore, SOD models trained with CELoss hold biased prior and tend to
predict unknown pixels as the background,
consequently leading to low-recall detections.
%
%
The F-measure~\cite{van1974foundation} is a more sophisticated
and comprehensive evaluation metric which combines
precision and recall into a single score
and automatically offsets the unbalance between positive/negative samples.

In this paper, we  provide a uniform formulation 
in both training and evaluation for SOD.
By directly taking the evaluation metric,
\emph{i.e.} the F-measure, as the optimization target,
we perform F-measure maximizing in an end-to-end manner.
To perform end-to-end learning,
we propose the \emph{relaxed F-measure} to overcome the in-differentiability
in the standard F-measure formulation.
The proposed loss function, named FLoss, is decomposable
w.r.t the posterior $\hat{Y}$ and
thus can be appended to the back of a CNN as supervision without effort.
We test the FLoss on several state-of-the-art SOD
architectures and witness a visible performance gain.
Furthermore, the proposed FLoss holds considerable gradients even in the saturated area,
resulting in polarized predictions that are stable against the threshold.
Our proposed FLoss enjoys three favorable properties:\vspace{-3pt}
\begin{itemize}[noitemsep]
  \item Threshold-free salient object detection.
  Models trained with FLoss produce contrastive saliency maps in which
  the foreground and background are clearly separated.
  Therefore,
  FLoss can achieve high performance under a wide range of threshold.

  \item Being able to deal with unbalanced data.
  Defined as the harmonic mean of precision and recall, the F-measure is able to establish a
  balance between samples of different classes.
  We experimentally evidence that our method can find a better compromise between
  precision and recall.

  \item Fast convergence. Our method quickly learns to focus on
  salient object areas after only hundreds of iterations,
  showing fast convergence speed.
\end{itemize}
\section{Related Work}
We review several CNN-based architectures for SOD and
the literature related to F-measure optimization.

\paragraph{Salient Object Detection (SOD).}
The convolutional neural network (CNN) is proven to be dominant in many sub-areas
of computer vision.
Significant progress has been achieved since the presence of CNN in SOD.
The DHS net~\cite{liu2016dhsnet} is one of the pioneers of using CNN for SOD.
DHS firstly produces a coarse saliency map with global cues, including contrast, objectness \etal.
Then the coarse map is progressively refined with a hierarchical recurrent CNN.
The emergence of the fully convolutional network (FCN)~\cite{long2015fully} provides an 
elegant way to perform the end-to-end pixel-wise inference.
DCL~\cite{li2016deep} uses a two-stream architecture to process contrast information 
in both pixel and patch levels.
The FCN-based sub-stream produces a saliency map with pixel-wise accuracy, 
and the other network stream performs inference on each object segment.
Finally, a fully connected CRF~\cite{krahenbuhl2011efficient} is used to 
combine the pixel-level and segment-level semantics.

Rooted from the HED~\cite{xie2015holistically} for edge detection,
aggregating multi-scale side-outputs is proven to be effective in refining dense  predictions
especially when the detailed local structures are required to be preserved.
In the HED-like architectures, deeper side-outputs capture rich semantics and
shallower side-outputs contain high-resolution details.
Combining these representations of different levels will lead to significant performance
improvements.
DSS~\cite{hou2017deeply} introduces deep-to-shallow short connections
across different side-outputs to refine the shallow side-outputs with deep semantic features.
The deep-to-shallow short connections enable the shallow side-outputs to
distinguish real salient objects from the background
and meanwhile retain the high resolution.
Liu \etal \cite{Liu2019PoolSal} design a pooling-based module to
efficiently fuse convolutional features from a top-down pathway.
The idea of imposing top-down refinement has also
been adopted in Amulet~\cite{zhang2017amulet},
and enhanced by Zhao \etal~\cite{zhao2018hifi} with bi-directional refinement.
Later, Wang \etal \cite{wang2018salient} propose a visual attention-driven model
that bridges the gap between SOD and eye fixation prediction.
These methods mentioned above tried to refine SOD by
introducing a more powerful network architecture, 
from recurrent refining network to multi-scale side-output fusing.
We refer the readers to a recent survey ~\cite{BorjiCVM2019} for more details.

\paragraph{F-measure Optimization.} 
Despite having been utilized as a common performance metric
in many application domains,
optimizing the F-measure doesn't draw much attention until very recently.
The works aiming at optimizing the F-measure can be divided into two subcategories
~\cite{dembczynski2013optimizing}:
(a) structured loss minimization methods such as~\cite{petterson2010reverse, petterson2011submodular}
which optimize the F-measure as the target during training;
and (b) plug-in rule approaches which optimize the F-measure during inference phase
~\cite{jansche2007maximum,dembczynski2011exact,quevedo2012multilabel,nan2012optimizing}.

Much of the attention has been drawn to the study of the latter subcategory:
finding an optimal threshold value which leads to a maximal F-measure given predicted
posterior $\hat{Y}$.
There are few articles about optimizing the F-measure during the training phase.
Petterson \etal ~\cite{petterson2010reverse} optimize the F-measure indirectly
by maximizing a loss function associated to the F-measure.
Then in their successive work~\cite{petterson2011submodular} they construct an
upper bound of the discrete F-measure
and then maximize the F-measure by optimizing its upper bound.
These previous studies either work as post-processing,
or are in-differentiable w.r.t posteriors,
making them hard to be applied to the deep learning framework.

\section{Optimizing the F-measure for SOD}

\subsection{The Relaxed F-measure}
In the standard F-measure, the true positive, 
false positive and false negative are defined as the number of corresponding samples:
\begin{equation}
\begin{split}
TP(\dot{Y}^t, Y) &= \sum\nolimits_i 1(y_i==1 \ \text{and} \ \dot{y}^t_i==1), \\
FP(\dot{Y}^t, Y) &= \sum\nolimits_i 1(y_i==0 \ \text{and} \ \dot{y}^t_i==1), \\
FN(\dot{Y}^t, Y) &= \sum\nolimits_i 1(y_i==1 \ \text{and} \ \dot{y}^t_i==0), \\
\end{split}
\label{eq:tpfp0}
\end{equation}
where $Y$ is the ground-truth, $\dot{Y}^t$ is the binary prediction binarized by threshold $t$
and $Y$ is the ground-truth saliency map.
$1(\cdot)$ is an indicator function that evaluates to $1$ if its argument is true and 0 otherwise.

To incorporate the F-measure into CNN and optimize it in an end-to-end manner,
we define a decomposable F-measure that is differentiable over posterior $\hat{Y}$.
Based on this motivation, we reformulate the true positive, false positive and false negative
based on the continuous posterior $\hat{Y}$:
\begin{equation}
\begin{split}
TP(\hat{Y}, Y) &= \sum\nolimits_i \hat{y}_i \cdot y_i, \\
FP(\hat{Y}, Y) &= \sum\nolimits_i \hat{y}_i \cdot (1 - y_i), \\
FN(\hat{Y}, Y) &= \sum\nolimits_i (1-\hat{y}_i) \cdot y_i \ . \\
\end{split}
\label{eq:tpfp}
\end{equation}
%
Given the definitions in Eq.~\ref{eq:tpfp}, precision $p$ and recall $r$ are:
\begin{equation}
p(\hat{Y}, Y) = \frac{TP}{TP + FP},\quad r(\hat{Y}, Y) = \frac{TP}{TP + FN}.
\label{pr}
\end{equation}
Finally, our \emph{relaxed F-measure} can be written as:
\begin{equation}
\begin{split}
F(\hat{Y}, Y) &= \frac{(1+\beta^2) p \cdot r}{\beta^2 p + r} ,\\
              &= \frac{(1 + \beta^2)TP}{\beta^2(TP + FN) + (TP + FP)} ,\\
              &= \frac{(1 + \beta^2)TP}{H},
\end{split}
\label{f}
\end{equation}
where $H\! =\! \beta^2(TP + FN) + (TP + FP)$.
Due to the relaxation in Eq.~\ref{eq:tpfp}, Eq.~\ref{f} is decomposable w.r.t the
posterior $\hat{Y}$, therefore can be integrated in CNN architecture trained with
back-prop.

\subsection{Maximizing F-measure in CNNs}
In order to maximize the \emph{relaxed F-measure} in CNNs in an end-to-end manner,
we define our proposed F-measure based loss (FLoss) function $\mathcal{L}_{F}$ as:
\begin{equation}
\mathcal{L}_{F}(\hat{Y}, Y) = 1 - F = 1 - \frac{(1 + \beta^2)TP}{H}\label{eq:floss}.
\end{equation}
Minimizing $\mathcal{L}_{F}(\hat{Y}, Y)$ is equivalent to maximizing the \emph{relaxed F-measure}.
Note again that $\mathcal{L}_{F}$ is calculated directly from the raw prediction $\hat{Y}$ without
thresholding.
Therefore, $\mathcal{L}_{F}$ is differentiable
over the prediction $\hat{Y}$ and can be plugged into CNNs.
The partial derivative of loss $\mathcal{L}_{F}$ over network activation $\hat{Y}$ at location $i$ is:
\begin{equation}
\begin{split}
\frac{\partial \mathcal{L}_{F}}{\partial \hat{y}_i}
  &= -\frac{\partial F}{\partial \hat{y}_i} \\
  &= -\Big(\frac{\partial F}{\partial TP}\cdot \frac{\partial TP}{\partial \hat{y}_i} +
        \frac{\partial F}{\partial H }\cdot \frac{\partial H }{\partial \hat{y}_i}\Big) \\
  &= -\Big(\frac{(1+\beta^2)y_i}{H} - \frac{(1+\beta^2)TP}{H^2}\Big) \\
  &= \frac{(1+\beta^2)TP}{H^2} - \frac{(1+\beta^2)y_i}{H} .\\
\end{split}\label{eq:grad-floss}
\end{equation}

There is another alternative to Eq.~\ref{eq:floss} which maximize the log-likelihood of F-measure:
\begin{equation}
  \mathcal{L}_{\log F}(\hat{Y}, Y) = -\log(F)\label{eq:logfloss},
\end{equation}
and the corresponding gradient is
\begin{equation}
\frac{\partial \mathcal{L}_{\log F}}{\partial \hat{y}_i} =
\frac{1}{F}\left[\frac{(1+\beta^2)TP}{H^2} - \frac{(1+\beta^2)y_i}{H}\right]. \\
\label{eq:grad-logfloss}
\end{equation}
We will theoretically and experimentally analyze the advantage of
FLoss against Log-FLoss and CELoss in terms of
producing polarized and high-contrast saliency maps.

\subsection{FLoss vs Cross-entropy Loss}\label{sec:cel-vs-floss}
To demonstrate the superiority of our  FLoss over the alternative Log-FLoss and
the \emph{cross-entropy loss} (CELoss),
we compare the definition, gradient and surface plots of these three loss functions.
The definition of CELoss is:
\begin{equation}
\mathcal{L}_{CE}(\hat{Y}, Y) \!=\! -\sum\nolimits_i^{|Y|} 
\left(y_i  \log{\hat{y}_i} + (1\!-\!y_i)  \log{(1\!-\!\hat{y}_i)}\right),
\label{eq:celoss}
\end{equation}
where $i$ is the spatial location of the input image and $|Y|$ is the number of pixels of the input image.
The gradient of $\mathcal{L}_{CE}$ w.r.t prediction $\hat{y}_i$ is:
\begin{equation}
\frac{\partial \mathcal{L}_{CE}}{\partial \hat{y}_i} = \frac{y_i}{\hat{y}_i} - \frac{1 - y_i}{1 - \hat{y}_i}.
\label{eq:grad-celoss}
\end{equation}

\CheckRmv{
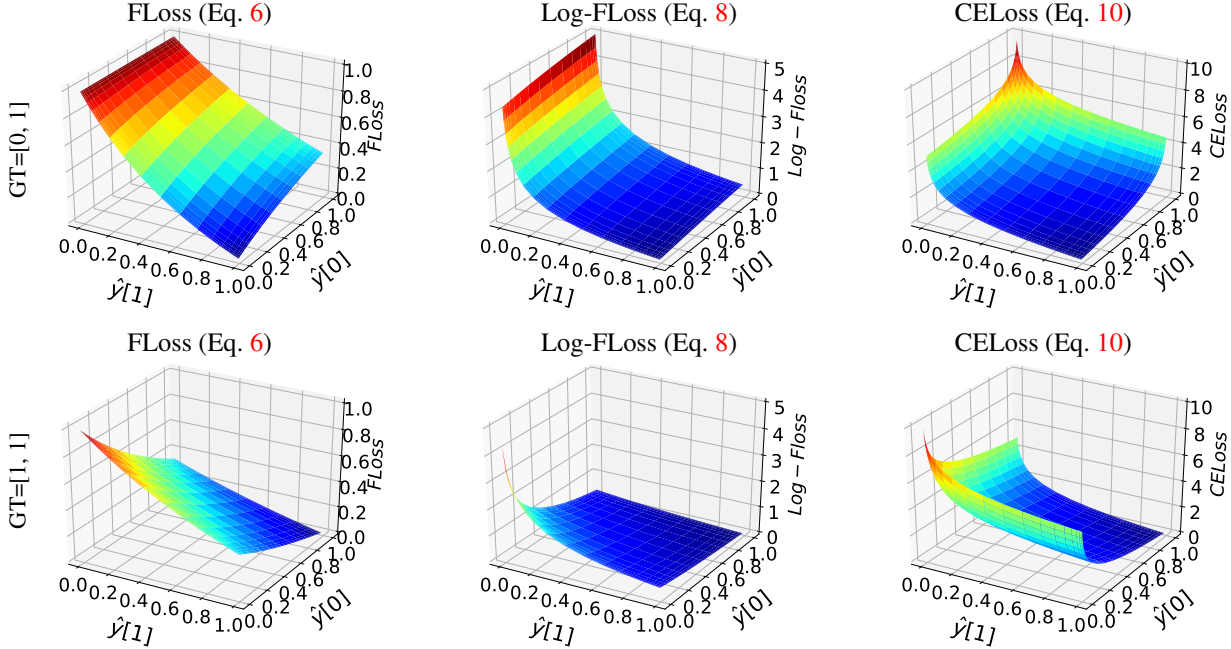
\begin{figure*}[!htb]
  \centering
  \begin{overpic}[width=0.9\linewidth]{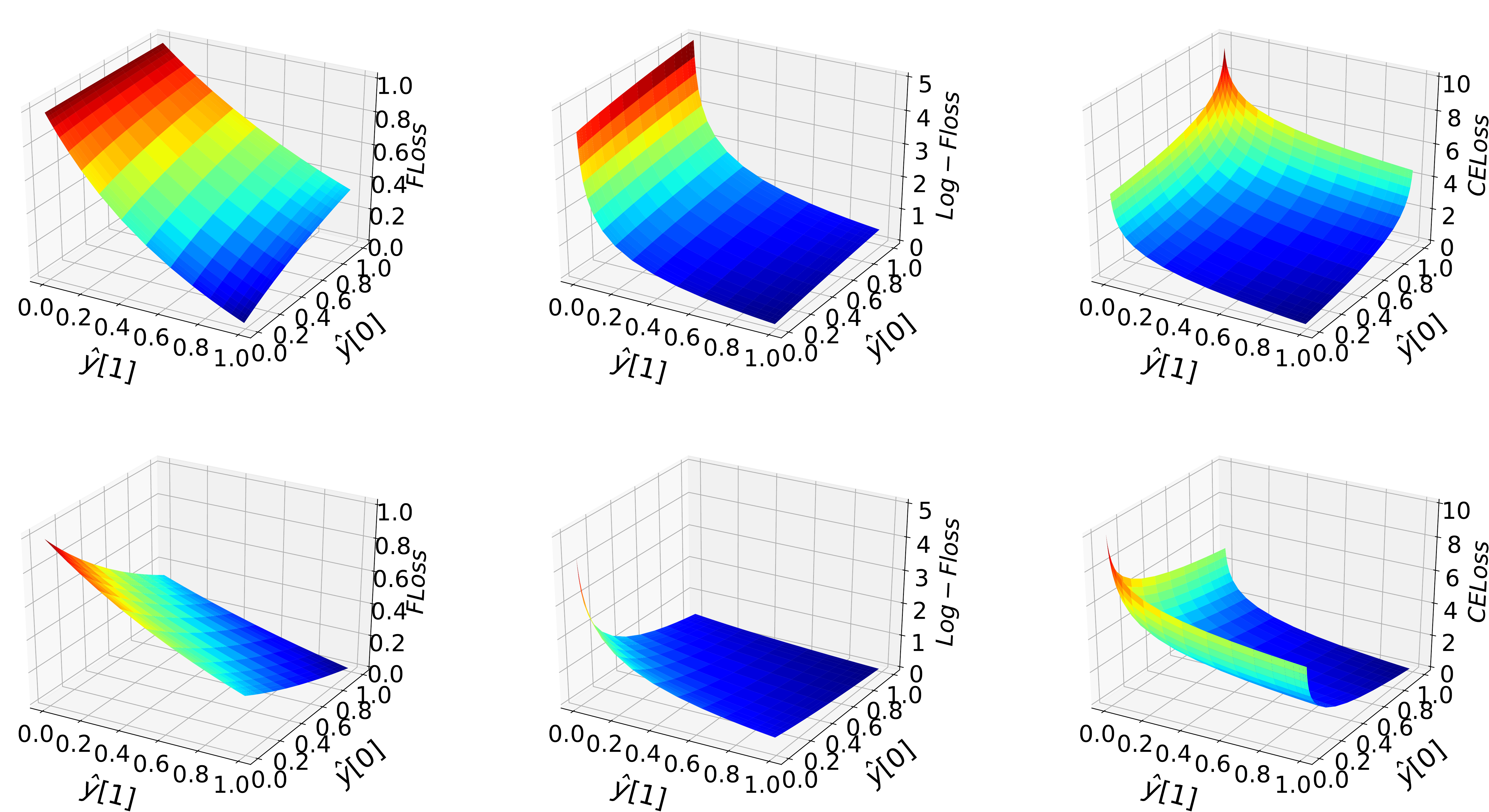}
  \put(7,53){FLoss (Eq.~\ref{eq:floss})}
  \put(42,53){Log-FLoss (Eq.~\ref{eq:logfloss})}
  \put(77,53){CELoss (Eq.~\ref{eq:celoss})}
  \put(7,25){FLoss (Eq.~\ref{eq:floss})}
  \put(42,25){Log-FLoss (Eq.~\ref{eq:logfloss})}
  \put(77,25){CELoss (Eq.~\ref{eq:celoss})}
  \put(-3,38){\rotatebox{90}{\small{GT=[0, 1]}}}
  \put(-3,10){\rotatebox{90}{\small{GT=[1, 1]}}}
  \end{overpic}\vspace{2pt}
  \caption{Surface plot of different loss functions in a 2-point 2-class classification circumstance.
  Columns from left to right: F-measure loss defined in Eq.~\ref{eq:floss},
  log F-measure loss defined in Eq.~\ref{eq:logfloss} and
  cross-entropy loss in Eq.~\ref{eq:celoss}.
  In top row the ground-truth is [0, 1] and in bottom row the ground-truth
  is [1, 1].
  Compared with cross-entropy loss and Log-FLoss,
  FLoss holds considerable gradient even in the saturated area, which will force
  to produce polarized predictions.
  }\label{fig:loss-surface}\vspace{-6pt}
\end{figure*}
}

As revealed in Eq.~\ref{eq:grad-floss} and Eq.~\ref{eq:grad-celoss}, the gradient of CELoss
$\frac{\partial \mathcal{L}_{CE}}{\partial \hat{y}_i}$ relies only on the
prediction/ground-truth of a single pixel $i$;
whereas in FLoss $\frac{\partial \mathcal{L}_{F}}{\partial \hat{y}_i}$
is globally determined by
the prediction and ground-truth of ALL pixels in the image.
We further compare the surface plots
of FLoss, Log-FLoss and CELoss in a two points binary classification problem.
The results are in Fig.~\ref{fig:loss-surface}.
The two spatial axes represent the prediction $\hat{y}_0$ and $\hat{y}_1$,
and the $z$ axis indicates the loss value.

As shown in Fig.~\ref{fig:loss-surface},
the gradient of FLoss is different from that of CELoss and Log-FLoss in two aspects:
(1) Limited gradient: the FLoss holds limited gradient values even
the predictions are far away from the ground-truth.
This is crucial for CNN training because it prevents the notorious gradient explosion problem.
Consequently, FLoss allows larger learning rates in the training phase, as evidenced by
our experiments.
(2) Considerable gradients in the saturated area: in CELoss, the gradient decays
when the prediction gets closer to the ground-truth,
while FLoss holds considerable gradients even in the saturated area.
This will force the network to have polarized predictions.
Salient detection examples in Fig.~\ref{fig:examples} illustrate the `high-contrast'
and polarized predictions.

\section{Experiments and Analysis}

\subsection{Experimental Configurations}
\textbf{Dataset and data augmentation.}
We uniformly train our model and competitors on the MSRA-B~\cite{liu2011learning}
training set for a fair comparison.
The MSRA-B dataset with 5000 images in total is equally split into training/testing
subsets.
We test the trained models on 5 other SOD datasets:
ECSSD~\cite{yan2013hierarchical},
HKU-IS~\cite{li2015visual},
PASCALS~\cite{li2014secrets},
SOD~\cite{movahedi2010design},
and DUT-OMRON~\cite{movahedi2010design}.
More statistics of these datasets are shown in Table~\ref{tab:dset-stats}.
It's worth mentioning that the challenging degree of a dataset is determined by many factors
such as the number of images, number of objects in one image, the contrast of salient object w.r.t the background,
the complexity of salient object structures, the center bias of salient objects and
the size variance of images \emph{etc}.
Analyzing these details is out of the scope of this paper,
we refer the readers to~\cite{dpfan2018soc} for more analysis of datasets.

\CheckRmv{
\begin{table}[!htb]
\centering
  \renewcommand{\arraystretch}{1.00}
  \setlength\tabcolsep{4pt}
  \resizebox{0.45\textwidth}{!}{
\begin{tabular}{r|c|c|c|c}
\toprule[1pt]
\textbf{Dataset~~~~} & \textbf{\#Images} &\textbf{Year} &\textbf{Pub.} & \textbf{Contrast} \\
\midrule[1pt]
MSRA-B~\cite{liu2011learning} & 5000 & 2011 & TPAMI & High\\
ECSSD~\cite{yan2013hierarchical}   & 1000 & 2013 & CVPR & High\\
HKU-IS~\cite{li2015visual} & 1447 & 2015 & CVPR& Low \\
PASCALS~\cite{li2014secrets}  & 850 & 2014 & CVPR & Medium\\
SOD~\cite{movahedi2010design} & 300  & 2010 & CVPRW & Low\\
DUT-OMRON~\cite{yang2013saliency} & 5168 & 2013 & CVPR & Low\\
\bottomrule[1pt]
\end{tabular}
}
\vspace{4pt}
\caption{Statistics of SOD datasets.
  `\#Images' indicates the number of images in a dataset
  and `contrast' represents the general contrast between foreground/background.
  The lower the contrast, the more challenging the dataset is.
}\label{tab:dset-stats}\vspace{-5pt}
\end{table}
}

Data augmentation is critical to generating sufficient data for training deep CNNs.
We fairly perform data augmentation for the original implementations and their FLoss variants.
For the DSS~\cite{hou2017deeply} and DHS~\cite{liu2016dhsnet} architectures we perform only
horizontal flip on both training images and saliency maps just as DSS did.
Amulet~\cite{zhang2017amulet} only allows $256\!\times\!256$ inputs.
We randomly crop/pad the original data to get square images, then resize them to meet the shape requirement.

\CheckRmv{
\begin{table*}[!htb]
  \centering
  \scriptsize
  \renewcommand{\arraystretch}{1.2}
  \renewcommand{\tabcolsep}{2pt}
  \resizebox{0.99\textwidth}{!}{
  \begin{tabular}{lcc|ccc|ccc|ccc|ccc|ccc}
  \toprule[1pt] &
  \multicolumn{2}{c}{Training data} &
  \multicolumn{3}{c}{ECSSD~\cite{yan2013hierarchical}} &
  \multicolumn{3}{c}{HKU-IS~\cite{li2015visual}} &
  \multicolumn{3}{c}{PASCALS~\cite{li2014secrets}} &
  \multicolumn{3}{c}{SOD~\cite{movahedi2010design}} &
  \multicolumn{3}{c}{DUT-OMRON~\cite{movahedi2010design}}\\
  \cmidrule(l){2-3} \cmidrule(l){4-6} \cmidrule(l){7-9} \cmidrule(l){10-13}
  \cmidrule(l){13-15} \cmidrule(l){16-18}
  Model & Train & \#Images & MaxF & MeanF & MAE & MaxF & MeanF & MAE &
          MaxF & MeanF & MAE & MaxF & MeanF & MAE & MaxF & MeanF & MAE \\
  \midrule[1pt]
  Log-FLoss&
  MB~\cite{liu2011learning} & 2.5K &
  .909 & .891 & .057 & .903 & .881 & .043 & .823 & .808 &
  .101 & .838 & .817 & .122 & .770 & .741 & .062 \\
  \textbf{FLoss}&
  MB~\cite{liu2011learning} & 2.5K &
  .914 & .903 & .050 & .908 & .896 & .038 & .829 & .818 &
  .091 & .843 & .838 & .111 & .777 & .755 & .067 \\
  \bottomrule[1pt]
  \vspace{0pt}
  \end{tabular}
  }
  \caption{Performance comparison of Log-FLoss (Eq.~\ref{eq:logfloss}) and FLoss (Eq.~\ref{eq:floss}).
  FLoss performs better than Log-FLoss on most datasets
  in terms of MaxF, MeanF and MAE.
  Specifically FLoss enjoys a large improvement in terms of MeanF because
  of its high-contrast predictions.
  }
  \label{tab:floss-vs-logfloss}
\end{table*}
}

\textbf{Network architecture and hyper-parameters.}
We test our proposed FLoss on 3 baseline methods:
Amulet~\cite{zhang2017amulet}, DHS~\cite{liu2011learning} and DSS~\cite{hou2017deeply}.
To verify the effectiveness of FLoss (Eq.~\ref{eq:floss}),
we replace the loss functions 
of the original implementations with FLoss
and keep all other configurations unchanged.
As explained in Sec.~\ref{sec:cel-vs-floss}, the FLoss allows a larger base learning rate due to
limited gradients.
We use the base learning rate $10^4$ times the original settings.
For example, in DSS the base learning rate is $10^{-8}$, while in our F-DSS, the base learning
rate is $10^{-4}$.
All other hyper-parameters are consistent with the original implementations for a fair comparison.

\textbf{Evaluation metrics.} We evaluate the performance of saliency maps
in terms of maximal F-measure (MaxF), mean F-measure (MeanF) and mean absolute error
(MAE = $\frac{1}{N}\sum_i^N |\hat{y}_i - y_i|$).
The factor $\beta^2$ in Eq.~\ref{eq:def-f} is set to 0.3 as suggested by
~\cite{achanta2009frequency, hou2017deeply, li2016deep,  liu2016dhsnet, wang2016saliency}.
By applying series thresholds $t\in \mathcal{T}$ to the saliency map $\hat{Y}$, we obtain
binarized saliency maps $\dot{Y}^t$ with different precisions, recalls and F-measures.

Then the optimal threshold $t_o$ is obtained by exhaustively searching the testing set:
\begin{equation}
t_o = \argmax_{t\in \mathcal{T}} F(Y, \dot{Y}^t).
\label{eq:optimal-t}
\end{equation}

Finally, we binarize the predictions with $t_o$ and evaluate the best F-measure:
\begin{equation}
  \text{MaxF} = F(Y, \dot{Y}^{t_o}),
  \label{eq:maxf}
\end{equation}
where $\dot{Y}^{t_o}$ is a binary saliency map binarized with $t_o$.
The MeanF is the average F-measure under different thresholds:
\begin{equation}
\text{MeanF} = \frac{1}{|\mathcal{T}|}\sum_{t\in \mathcal{T}} F(Y, \dot{Y}^t),
\label{eq:meanf}
\end{equation}
where $\mathcal{T}$ is the collection of possible thresholds.

\subsection{Log-FLoss vs FLoss}
Firstly we compare FLoss with its alternative, namely Log-FLoss defined in Eq.~\ref{eq:logfloss},
to justify our choice.
As analyzed in Sec.~\ref{sec:cel-vs-floss}, FLoss enjoys the advantage of having large gradients
in the saturated area that cross-entropy loss and Log-FLoss don't have.

\CheckRmv{
\begin{figure}[b]
  \centering
  \begin{overpic}[width=1\linewidth]{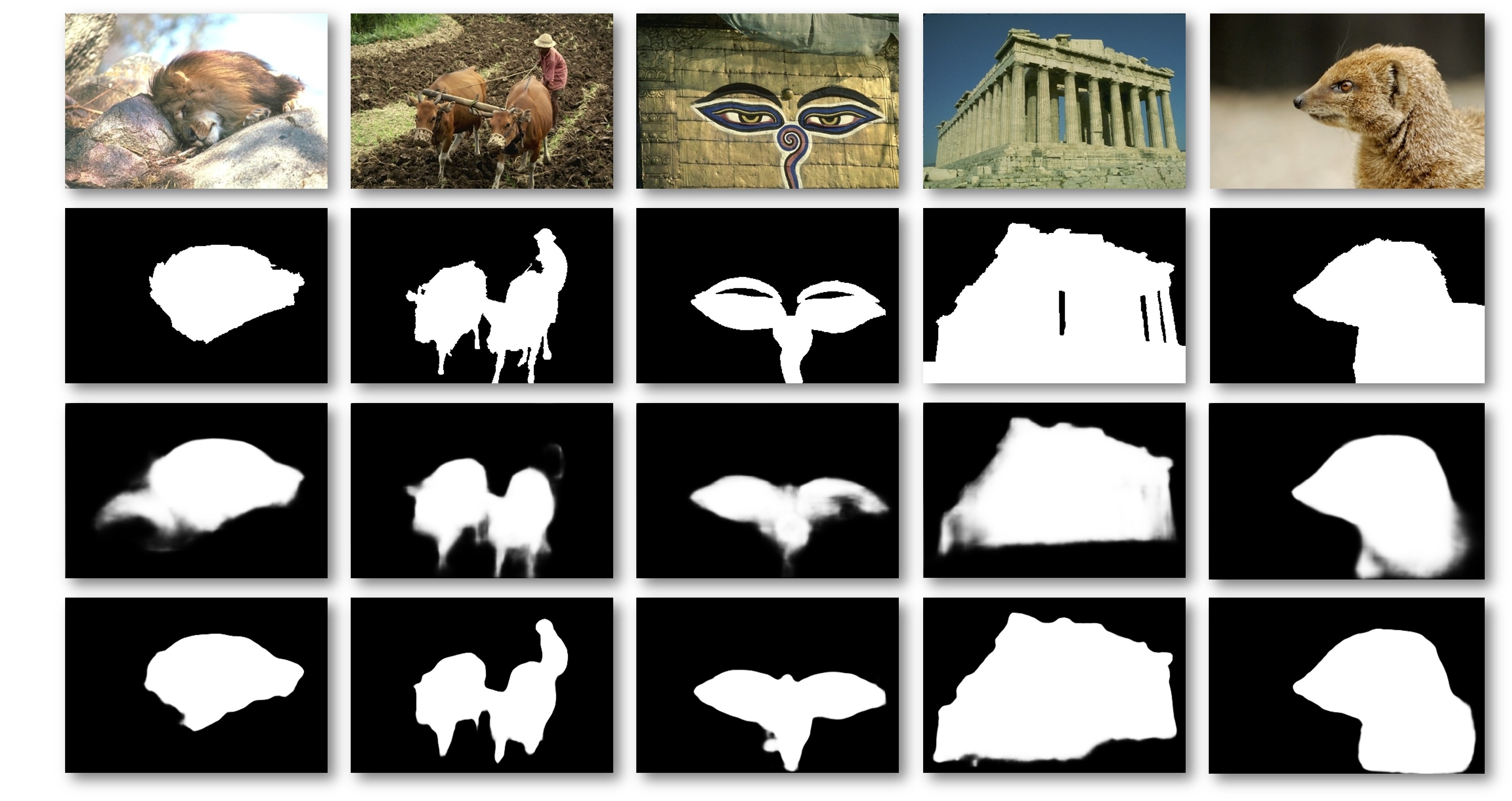}
  \put(-1,41){\rotatebox{90}{Image}}
  \put(-1,30){\rotatebox{90}{GT}}
  \put(-1,14){\rotatebox{90}{\footnotesize{Log-FLoss}}}
  \put(-1,2){\rotatebox{90}{\small{FLoss}}}
  \end{overpic}
  \caption{Example saliency maps by FLoss (bottom) and Log-FLoss (middle).
  Our proposed FLoss method produces high-contrast saliency maps.
  }\label{fig:examples-floss-vs-logfloss}\vspace{0pt}
\end{figure}
}

\CheckRmv{
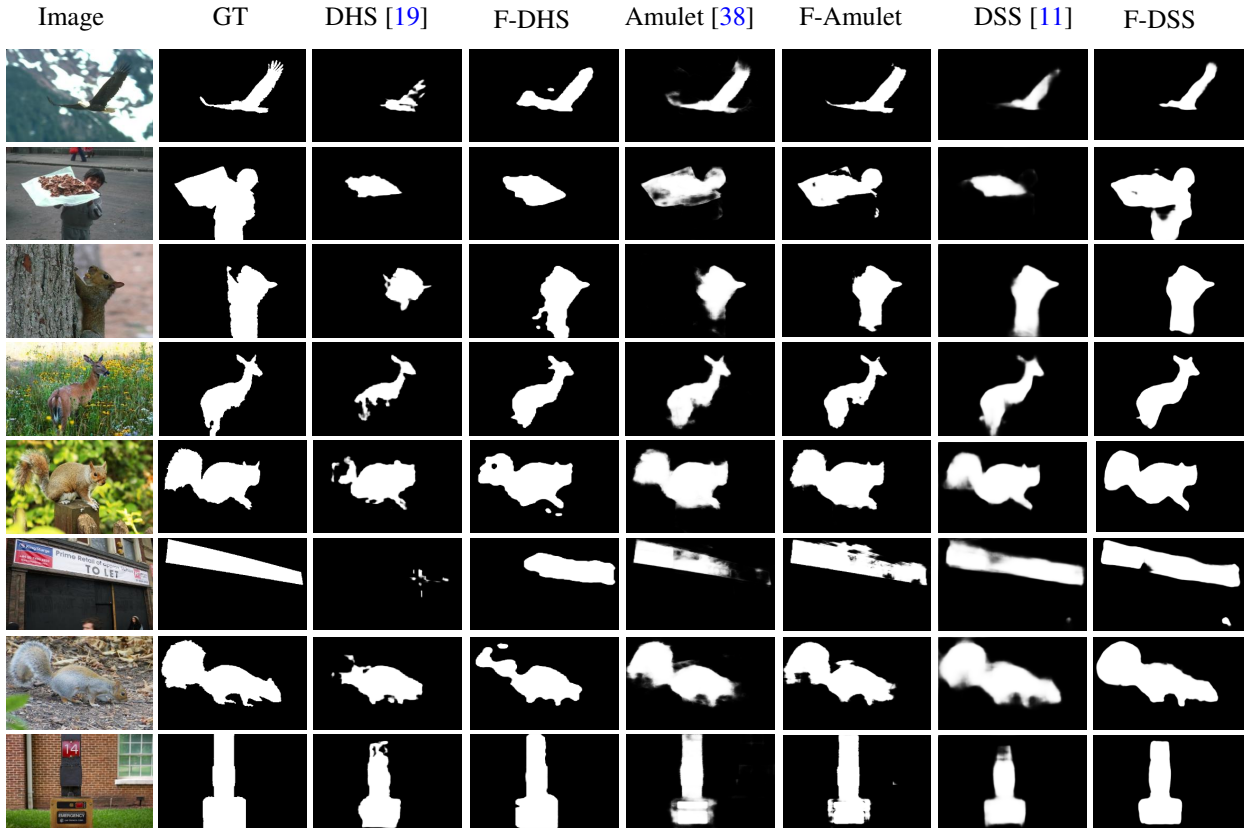
\begin{figure*}[!thb]
  \centering
  \begin{overpic}[width=0.95\linewidth]{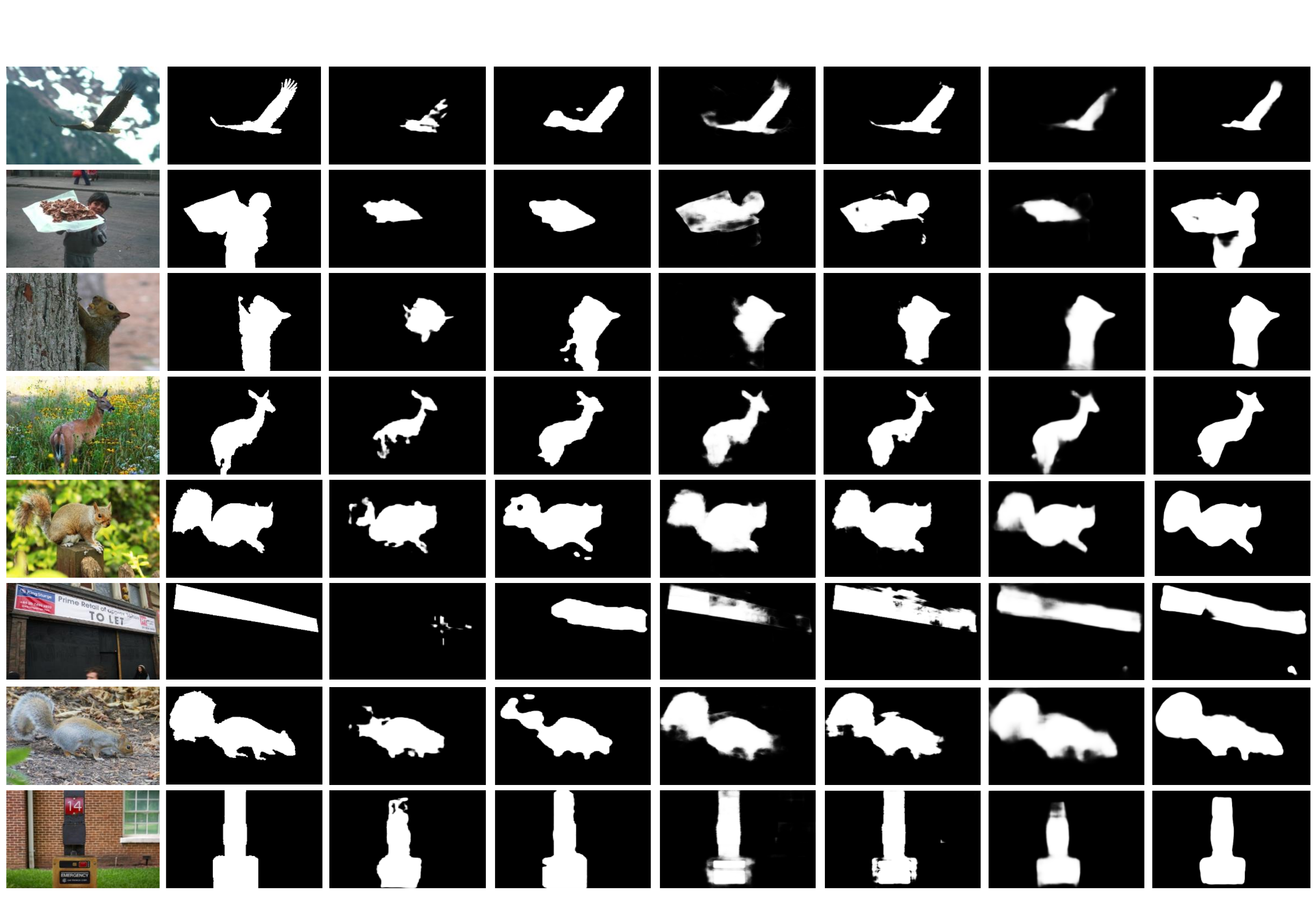}
  \put(3,65){Image}
  \put(17,65){GT}
  \put(26,65){DHS~\cite{liu2016dhsnet}}
  \put(39.5,65){\parbox{.5in}{F-DHS}}
  \put(50,65){Amulet~\cite{zhang2017amulet}}
  \put(64,65){F-Amulet}
  \put(78,65){DSS~\cite{hou2017deeply}}
  \put(90,65){\parbox{.5in}{F-DSS}}
  \end{overpic}
  \caption{Salient object detection examples on several popular datasets.
  F-DHS, F-Amulet and F-DSS indicate the original architectures trained with
our proposed FLoss. FLoss leads to sharp salient confidence, especially
  on the object boundaries.}\label{fig:examples}
\end{figure*}
}

To experimentally verify our assumption that FLoss will produce high-contrast predictions,
we train the DSS~\cite{hou2017deeply} model with FLoss and Log-FLoss, respectively.
The training data is MSRA-B~\cite{liu2011learning} and
hyper-parameters are kept unchanged with the original implementation, except for the base learning rate.
We adjust the base learning rate to $10^{-4}$ since our method accept larger learning rate, as explained
in Sec.~\ref{sec:cel-vs-floss}.
Quantitative results are in Table~\ref{tab:floss-vs-logfloss} and some example detected saliency maps
are shown in Fig.~\ref{fig:examples-floss-vs-logfloss}.

Although both of Log-FLoss
and FLoss use F-measure as maximization target,
FLoss derives polarized predictions with high foreground-background contrast,
as shown in Fig.~\ref{fig:examples-floss-vs-logfloss}.
The same conclusion can be drawn from Table~\ref{tab:floss-vs-logfloss} where
FLoss achieves higher Mean F-measure.
Which reveals that FLoss achieves higher
F-measure score under a wide range of thresholds.

\subsection{Evaluation results on open Benchmarks}

\CheckRmv{
\begin{table*}[!htp]
  \centering
  \scriptsize
  \renewcommand{\arraystretch}{1.5}
  \renewcommand{\tabcolsep}{2pt}
  \resizebox{0.99\textwidth}{!}{
  \begin{tabular}{lcc|ccc|ccc|ccc|ccc|ccc}
  \toprule[1pt] &
  \multicolumn{2}{c}{Training data} &
  \multicolumn{3}{c}{ECSSD~\cite{yan2013hierarchical}} &
  \multicolumn{3}{c}{HKU-IS~\cite{li2015visual}} &
  \multicolumn{3}{c}{PASCALS~\cite{li2014secrets}} &
  \multicolumn{3}{c}{SOD~\cite{movahedi2010design}} &
  \multicolumn{3}{c}{DUT-OMRON~\cite{movahedi2010design}}\\
  \cmidrule(l){2-3} \cmidrule(l){4-6} \cmidrule(l){7-9} \cmidrule(l){10-13} \cmidrule(l){13-15} \cmidrule(l){16-18}
  Model & Train & \#Images & MaxF & MeanF & MAE & MaxF & MeanF & MAE &
          MaxF & MeanF & MAE & MaxF & MeanF & MAE & MaxF & MeanF & MAE \\
  \midrule[1pt]
  \textbf{RFCN}~\cite{wang2016saliency} &
  MK~\cite{cheng2015global} &
  10K & .898 & .842 & .095 & .895 & .830 & .078 &
  .829 & .784 & .118 & .807 & .748 & .161 & - & - & -\\
  \textbf{DCL}~\cite{li2016deep} &
  MB~\cite{liu2011learning} &
  2.5K & .897 & .847 & .077 & .893 & .837 & .063 &
  .807 & .761 & .115 & .833 & .780 & .131 & .733 & .690 & .095 \\
  \textbf{DHS}~\cite{liu2016dhsnet} &
  MK~\cite{cheng2015global}+D~\cite{movahedi2010design} &
  9.5K & .905 & .876 & .066 & .891 & .860 & .059 &
  .820 & .794 & .101 & .819 & .793 & .136 & - & - & - \\
  \textbf{Amulet}~\cite{zhang2017amulet} & MK~\cite{cheng2015global} &
  10K & .912 & .898 & .059 & .889 & .873 & .052 &
  .828 & .813 & .092 & .801 & .780 & .146 & .737 & .719 & .083 \\
  \hline
  \textbf{DHS}~\cite{liu2016dhsnet} &
  MB &
  2.5K & .874 & .867 & .074 & .835 & .829 & .071 &
  .782 & .777 & .114 & .800 & .789 & .140 & .704 & .696 & \textbf{.078} \\
  \textbf{DHS+FLoss}~\cite{liu2016dhsnet} &
  MB & 2.5K &
  \textbf{.884} & \textbf{.879} & \textbf{.067} & \textbf{.859} &
  \textbf{.854} & \textbf{.061} & \textbf{.792} & \textbf{.786} &
  \textbf{.107} & \textbf{.801} & \textbf{.795} & \textbf{.138} &
  \textbf{.707} & \textbf{.701} & .079 \\
  \hline
  \textbf{Amulet}~\cite{zhang2017amulet} &
  MB & 2.5K &
  .881 & .857 & .076 & .868 & .837 & .061 &
  .775 & .753 & .125 & .791 & .776 & .149 & .704 & .663 & .098 \\
  \textbf{Amulet-FLoss}  &  MB & 2.5K &
  \textbf{.894} & \textbf{.883} & \textbf{.063} & \textbf{.880} &
  \textbf{.866} & \textbf{.051} & \textbf{.791} & \textbf{.776} &
  \textbf{.115} & \textbf{.805} & \textbf{.800} & \textbf{.138} &
  \textbf{.729} & \textbf{.696} & \textbf{.097} \\
  \hline
  \textbf{DSS}~\cite{hou2017deeply} & MB &
  2.5K & .908 & .889 & .060 & .899 & .877 & .048 &
  .824 & .806 & .099 & .835 & .815 & .125 & .761 & .738 & .071 \\
  \textbf{DSS+FLoss} &
  MB & 2.5K &
  \textbf{.914} & \textbf{.903} & \textbf{.050} & \textbf{.908} &
  \textbf{.896} & \textbf{.038} & \textbf{.829} & \textbf{.818} &
  \textbf{.091} & \textbf{.843} & \textbf{.838} & \textbf{.111} &
  \textbf{.777} & \textbf{.755} & \textbf{.067} \\
  \bottomrule[1pt]
  \vspace{0.5pt}
  \end{tabular}
  }\vspace{-8pt}
  \caption{Quantitative comparison of different methods on 6 popular datasets.
  Our proposed FLoss consistently improves performance in terms of both MAE (the smaller the better)
  and F-measure (the larger the better).
  Especially in terms of Mean F-measure, we outperform the state-of-the-art with very
  clear margins,  because our method is able to produce high-contrast predictions that can
  achieve high F-measure under a wide range of thresholds.
  }
  \label{tab:quantitative}\vspace{-12pt}
\end{table*}
}

We compare the proposed method with several baselines on 5 popular datasets.
Some example detection results are shown in Fig.~\ref{fig:examples} and
comprehensive quantitative comparisons are in Table~\ref{tab:quantitative}.
In general, FLoss-based methods can obtain considerable improvements compared with
their cross-entropy loss (CELoss) based counterparts
especially in terms of mean F-measure and MAE.
This is mainly because our method is stable against the threshold, leading to
high-performance saliency maps under a wide threshold range.
In our detected saliency maps, the foreground (salient objects) and background are well separated,
as shown in Fig.~\ref{fig:examples} and explained in Sec.~\ref{sec:cel-vs-floss}.

\subsection{Threshold Free Salient Object Detection}\label{sec:thres-free}

State-of-the-art SOD methods \cite{hou2017deeply,li2016deep,liu2016dhsnet,zhang2017amulet} 
often evaluate maximal F-measure as follows:
(a) Obtain the saliency maps $\hat{Y}_i$ with pretrained model;
(b) Tune the best threshold $t_o$ by exhaustive search on the testing set (Eq.~\ref{eq:optimal-t})
and binarize the predictions with $t_o$;
(c) Evaluate the maximal F-measure according to Eq.~\ref{eq:maxf}.
%
%
%
%

There is an obvious flaw in the above procedure:
the optimal threshold is obtained via an exhaustive search on the testing set.
%
Such procedure is impractical for real-world applications as we would 
not have annotated testing data.
And even if we tuned the optimal threshold on one dataset, it can not be widely applied
to other datasets.

\CheckRmv{
\begin{figure*}[t]
  \centering
    \begin{tabular}{@{}cc@{}}
      \begin{overpic}[width=.40\textwidth]{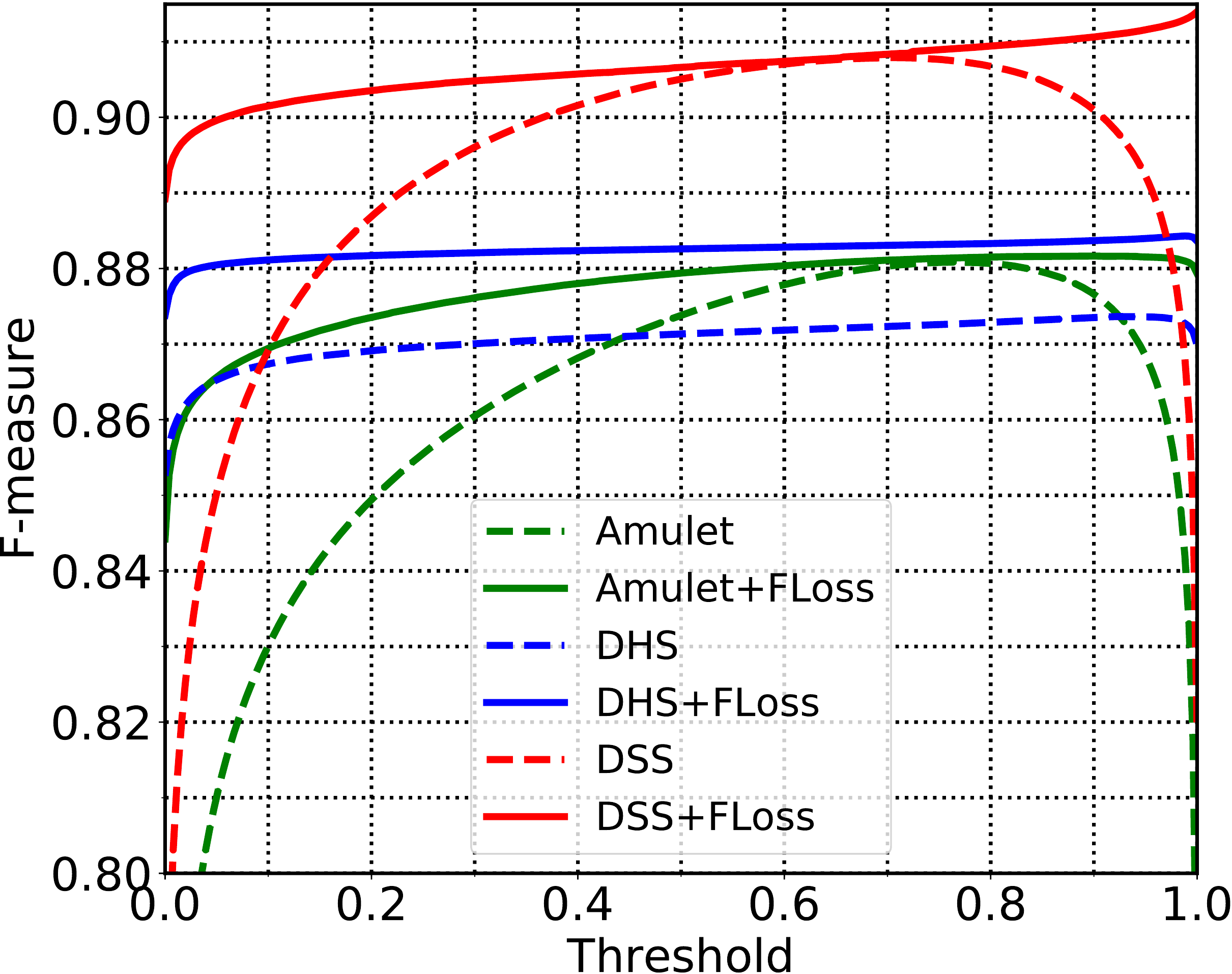}
      \put(60, 35.5){\scriptsize{\cite{zhang2017amulet}}} 
      \put(56, 26.5){\scriptsize{\cite{liu2016dhsnet}}} 
      \put(55, 17.5){\scriptsize{\cite{hou2017deeply}}} 
      \put(50, -5){(a)}
      \end{overpic}  &
      \begin{overpic}[width=.40\textwidth]{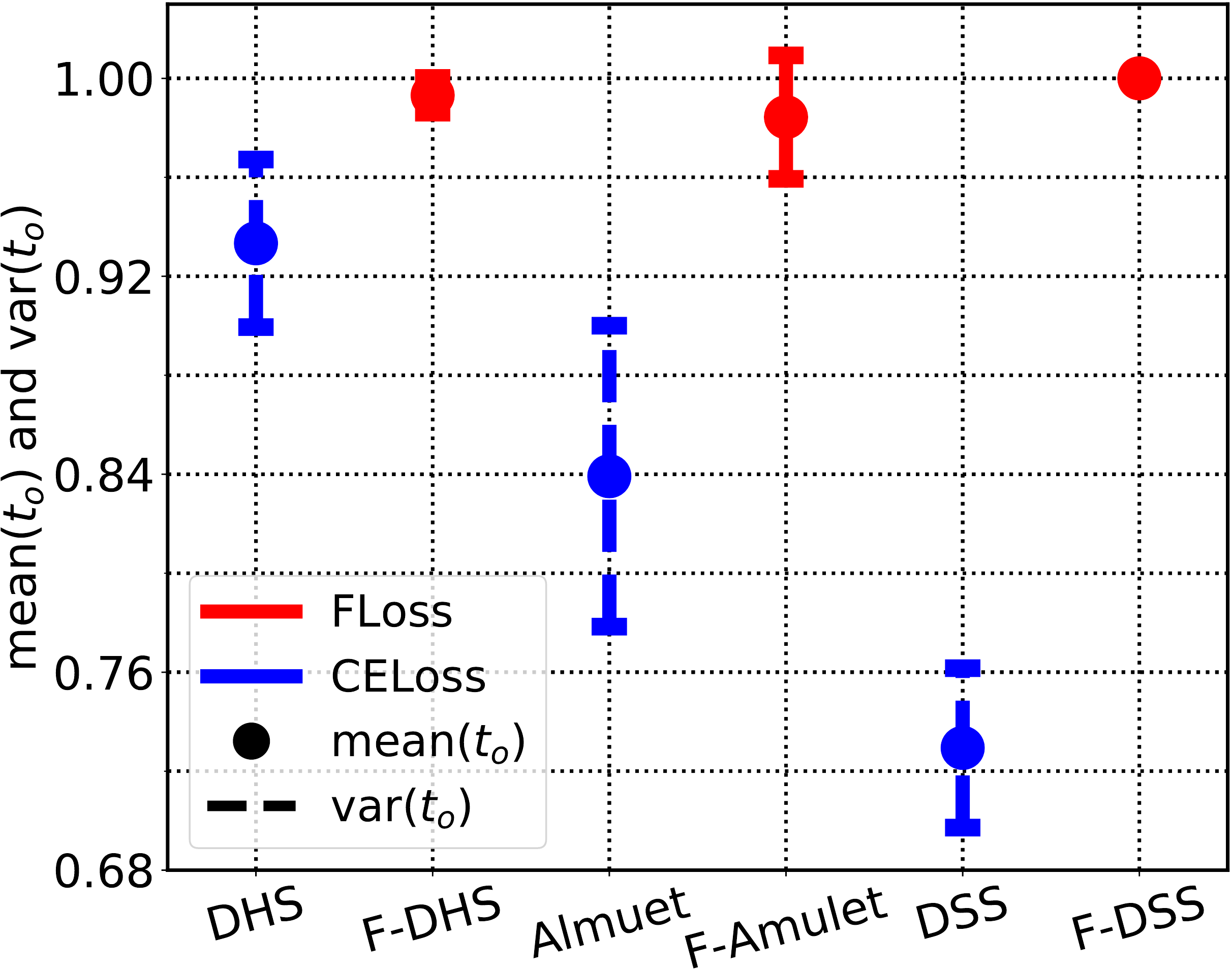}
        \put(50, -5){(b)}
      \end{overpic} \\
    \end{tabular}\vspace{10pt}
    \caption{(a) F-measures under different thresholds on the ECSSD dataset.
    (b) The mean and variance of optimal threshold $t_o$.
    FLoss-based methods hold stable $t_o$ across different datasets (lower $t_o$ variances) and different backbone
    architectures (F-DHS, F-Amulet and F-DSS hold very close mean $t_o$).
    }
    \label{fig:thres-free}
\end{figure*}
}

We further analyze the sensitivity of methods against thresholds in two aspects:
(1) model performance under different thresholds, which reflects the stability of
a method against threshold change,
(2) the mean and variance of optimal threshold $t_o$ on different datasets,
which represent the generalization ability of $t_o$ tuned on one dataset to others.

Fig.~\ref{fig:thres-free} (a) illustrates the F-measure w.r.t different thresholds.
For most methods without FLoss, the F-measure changes sharply with the threshold,
and the maximal F-measure (MaxF) presents only in a narrow threshold span.
While FLoss based methods are almost immune from the change of threshold.

Fig.~\ref{fig:thres-free} (b) reflects the mean and variance of $t_o$
across different datasets.
Conventional methods (DHS, DSS, Amulet) present unstable $t_o$ on different datasets,
as evidenced by their large variances.
While the $t_o$ of FLoss-based methods (F-DHS, F-Amulet, F-DSS)
stay unchanged across different datasets and different backbone network architectures.

In conclusion, the proposed FLoss is stable against threshold $t$
in three aspects:
(1) it achieves high performance under a wide range of threshold;
(2) optimal threshold $t_o$ tuned on one dataset can be transferred to others,
because $t_o$ varies slightly across different datasets;
and (3) $t_o$ obtained from one backbone architecture can be applied to other architectures.

\subsection{The Label-unbalancing Problem in SOD}

\CheckRmv{
\begin{figure}
  \centering
  \includegraphics[width=0.75\linewidth]{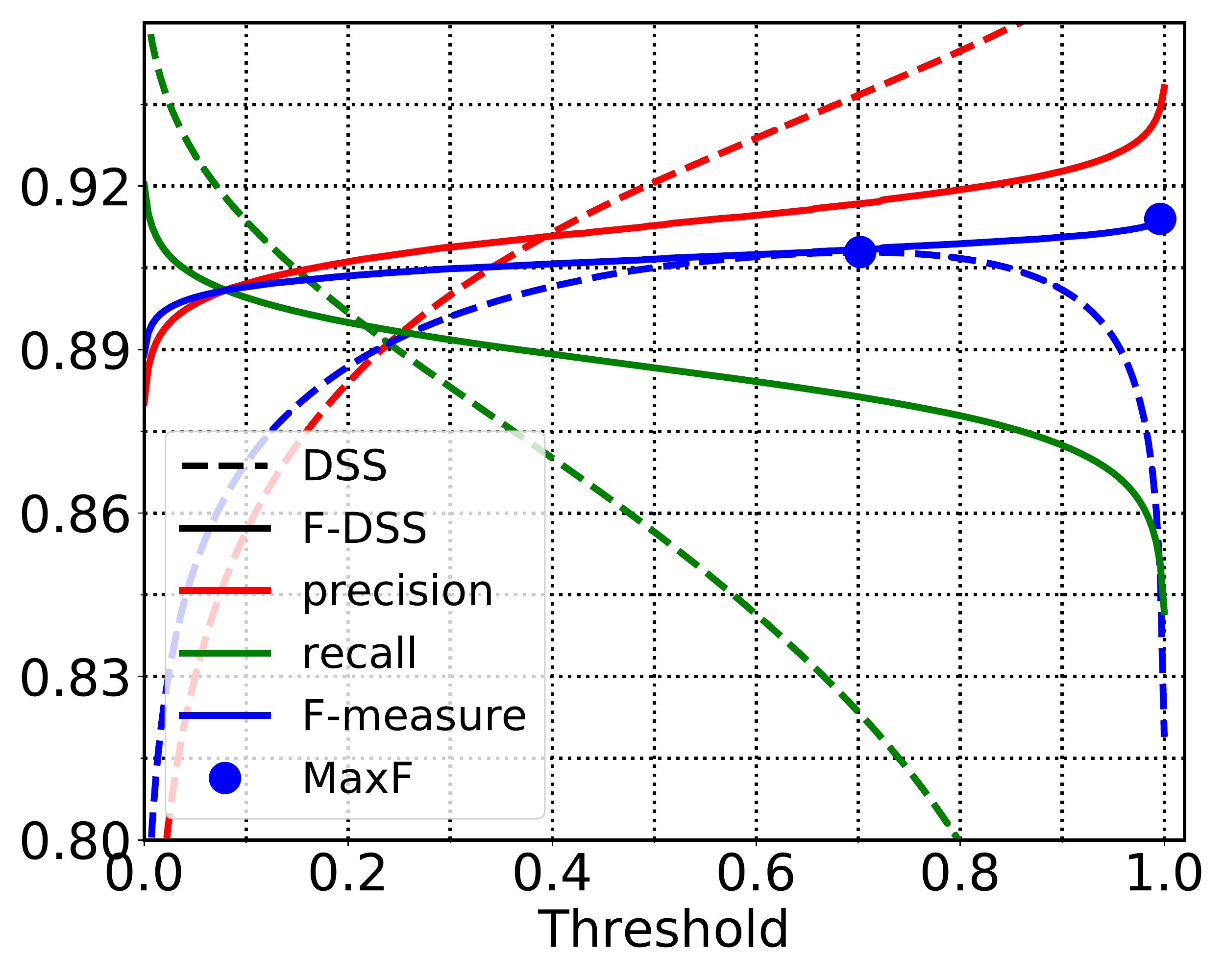}
  \caption{{\color{red}{\textbf{Precision}}}, {\color{green}{\textbf{Recall}}},
    {\color{blue}{\textbf{F-measure}}} and Maximal F-measure ({\color{blue}{$\bullet$}})  of
    DSS (\textbf{- - -})
    and F-DSS (\textbf{---}) under different thresholds.
    DSS tends to predict unknown pixels as the
    majority class--the background, resulting in high precision but low recall.
    FLoss is able to find a better compromise between precision and recall.
  }\label{fig:prf-thres}
\end{figure}
}

The foreground and background are biased in SOD where
most pixels belong to the non-salient regions.
The unbalanced training data will lead the model to local minimal
that tends to predict unknown pixels as the background.
Consequently, the recall will become a bottleneck to the performance during evaluations,
as illustrated in Fig.~\ref{fig:prf-thres}.

Although assigning loss weight to the positive/negative samples is a simple way to
offset the unbalancing problem, an additional experiment in Table~\ref{tab:balance}
reveals that our method performs better than simply assigning loss weight.
We define the \emph{balanced cross-entropy loss} with weight factor between 
positive/negative samples:
\begin{equation}
  \begin{split}
  \mathcal{L}_{balance} = \sum\nolimits_i^{|Y|} &w_1 \cdot y_i\log{\hat{y_i}} + \\
                          &w_0 \cdot (1-y_i)\log{(1-\hat{y_i})}.
  \end{split}
  \label{eq:balance-cross-entropy}
\end{equation}
The loss weights for positive/negative samples are determined by the 
positive/negative proportion in a mini-batch:
$w_1 = \frac{1}{|Y|}\sum_i^{|Y|} 1(y_i\!==\!0)$ and $w_0 = \frac{1}{|Y|}\sum_i^{|Y|} 1(y_i\!==\!1)$,
as suggested in~\cite{xie2015holistically} and \cite{shen2015deepcontour}.

\subsection{The Compromise Between Precision and Recall}

\CheckRmv{
\begin{table*}[t]
  \centering
  \scriptsize
  \renewcommand{\arraystretch}{1.2}
  \renewcommand{\tabcolsep}{2pt}
  \resizebox{0.99\textwidth}{!}{
  \begin{tabular}{lcc|ccc|ccc|ccc|ccc|ccc}
  \toprule[1pt] &
  \multicolumn{2}{c}{Training data} &
  \multicolumn{3}{c}{ECSSD~\cite{yan2013hierarchical}} &
  \multicolumn{3}{c}{HKU-IS~\cite{li2015visual}} &
  \multicolumn{3}{c}{PASCALS~\cite{li2014secrets}} &
  \multicolumn{3}{c}{SOD~\cite{movahedi2010design}} &
  \multicolumn{3}{c}{DUT-OMRON~\cite{movahedi2010design}}\\
  \cmidrule(l){2-3} \cmidrule(l){4-6} \cmidrule(l){7-9} \cmidrule(l){10-13}
  \cmidrule(l){13-15} \cmidrule(l){16-18}
  Model & Train & \#Images & MaxF & MeanF & MAE & MaxF & MeanF & MAE &
          MaxF & MeanF & MAE & MaxF & MeanF & MAE & MaxF & MeanF & MAE \\
  \midrule[1pt]
  \textbf{DSS}~\cite{hou2017deeply} &
  MB~\cite{liu2011learning} & 2.5K &
  .908 & .889 & .060 & .899 & .877 & .048 &
  .824 & .806 & .099 & .835 & .815 & .125 & .761 & .738 & .071 \\
  \textbf{DSS+Balance} &
  MB~\cite{liu2011learning} & 2.5K &
  .910 & .890 & .059 & .900 & .877 & .048 & .827 &
  .807 & .097 & .837 & .816 & .124 & .765 & .741 & .069 \\
  \textbf{DSS+FLoss} & MB~\cite{liu2011learning} & 2.5K &
  \textbf{.914} & \textbf{.903} & \textbf{.050} & \textbf{.908} & \textbf{.896} & \textbf{.038}
   & \textbf{.829} &
   \textbf{.818} & \textbf{.091} & \textbf{.843} & \textbf{.838} &
   \textbf{.111} & \textbf{.777} & \textbf{.755} & \textbf{.067} \\
  \bottomrule[1pt]
  \vspace{0pt}
  \end{tabular}
  }
  \vspace{-2pt}
  \caption{Performance comparisons across the original cross-entropy loss (Eq.~\ref{eq:celoss}),
  balanced cross-entropy loss (Eq.~\ref{eq:balance-cross-entropy}) and
  our proposed FLoss (Eq.~\ref{eq:floss}).
  Original cross-entropy learns a biased prior towards the major class (the background).
  This is evidenced by the low recall: many positive points
  are mis-predicted as negative because of biased prior.
  By assigning loss weights on foreground/background samples,
  the \emph{balanced cross-entropy loss} can alleviate the unbalancing problem.
  Our proposed method performs better than the \emph{balanced cross-entropy loss},
  because the F-measure criterion can automatically adjust data unbalance.
  }
    \vspace{-5pt}
  \label{tab:balance}
\end{table*}
}

Recall and precision are two conflict metrics.
In some applications, we care recall more than precision,
while in other tasks precision may be more important than recall.
The $\beta^2$ in Eq.~\ref{eq:def-f} balances the bias between precision and precision
when evaluating the performance of specific tasks.
For example, recent studies on edge detection
use~\cite{bsds500, xie2015holistically, shen2015deepcontour} $\beta^2=1$,
indicating its equal consideration on precision and recall.
While saliency detection~\cite{achanta2009frequency, hou2017deeply, li2016deep, liu2016dhsnet, wang2016saliency} usually uses $\beta^2=0.3$
to emphasize the precision over the recall.

As an optimization target, the FLoss should also be able to 
balance the favor between precision and recall.
We train models with different $\beta^2$ and comprehensively evaluate 
their performances in terms of precision, recall and F-measure.
Results in Fig.~\ref{fig:pr-beta} reveal that $\beta^2$ is a bias adjuster 
between precision and recall: 
larger $\beta^2$ leads to higher recall while lower $\beta^2$ results in higher precision.

\CheckRmv{
\begin{figure}[!htb]
  \centering
  \includegraphics[width=0.75\linewidth]{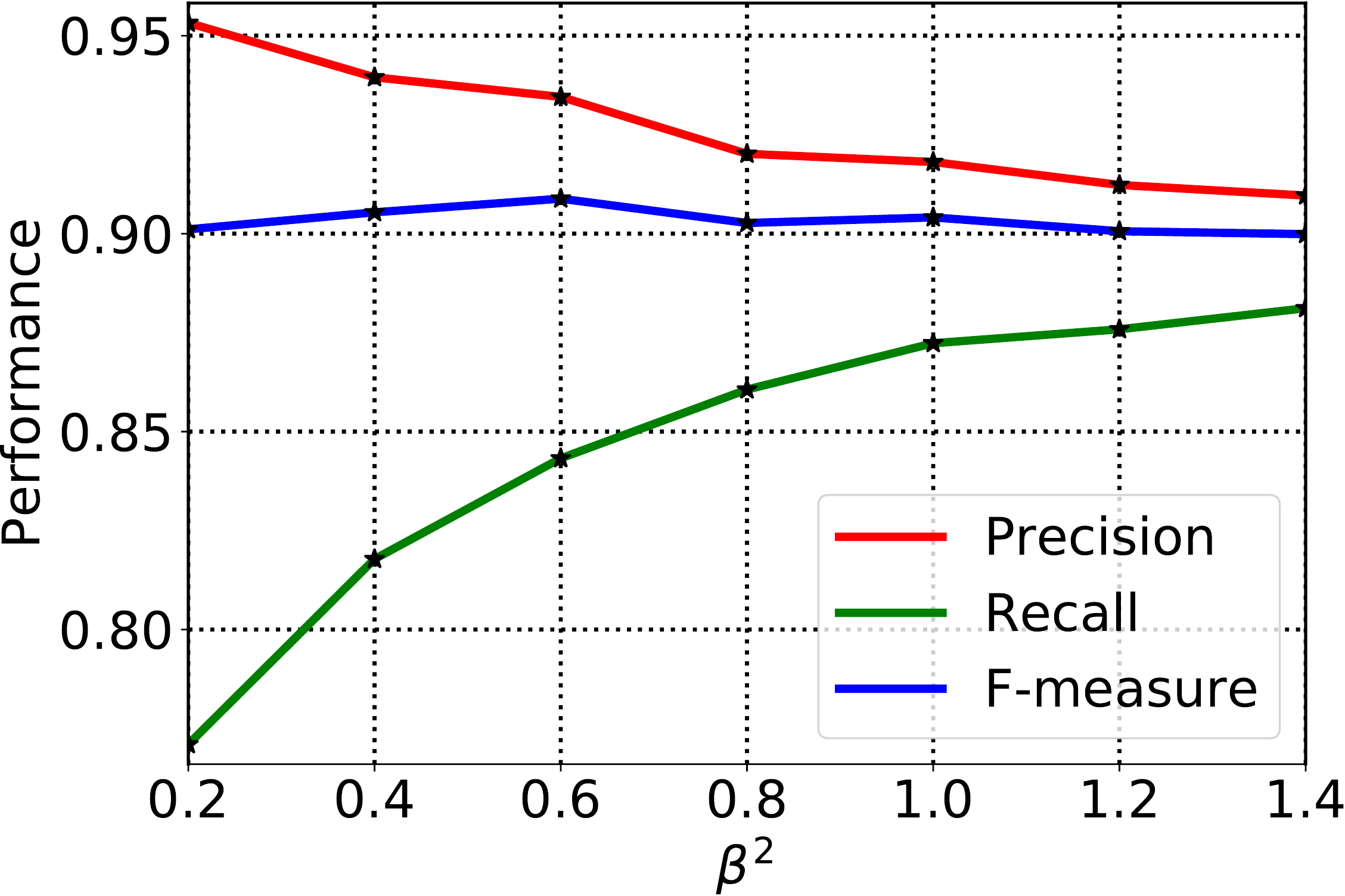}
  \caption{{\color{red}{\textbf{Precision}}}, {\color{green}{\textbf{Recall}}},
    {\color{blue}{\textbf{F-measure}}} of model trained under different $\beta^2$
    (Eq.~\ref{eq:def-f}).
    The precision decreases with the growing of $\beta^2$ whereas recall increases.
    This characteristic gives us much flexibility to adjust the balance 
    between recall and precision:
    use larger $\beta^2$ in a recall-first application and lower $\beta^2$ 
    otherwise.
  }\label{fig:pr-beta}\vspace{-12pt}
\end{figure}
}

\subsection{Faster Convergence and Better Performance}
In this experiment, we train three state-of-the-art saliency detectors (Amulet~\cite{zhang2017amulet}, 
DHS~\cite{liu2011learning} and DSS~\cite{hou2017deeply}) and their FLoss counterparts.
Then we plot the performance of all the methods at each checkpoint
to determine the converge speed and converged performance of respective models.
All the models are trained on the MB~\cite{liu2011learning} dataset
and tested on the ECSSD~\cite{yan2013hierarchical} dataset.
The results are shown in Fig.\ref{fig:f-iter}.

\CheckRmv{
\begin{figure}[!h]
  \centering
  \begin{overpic}[width=0.85\linewidth]{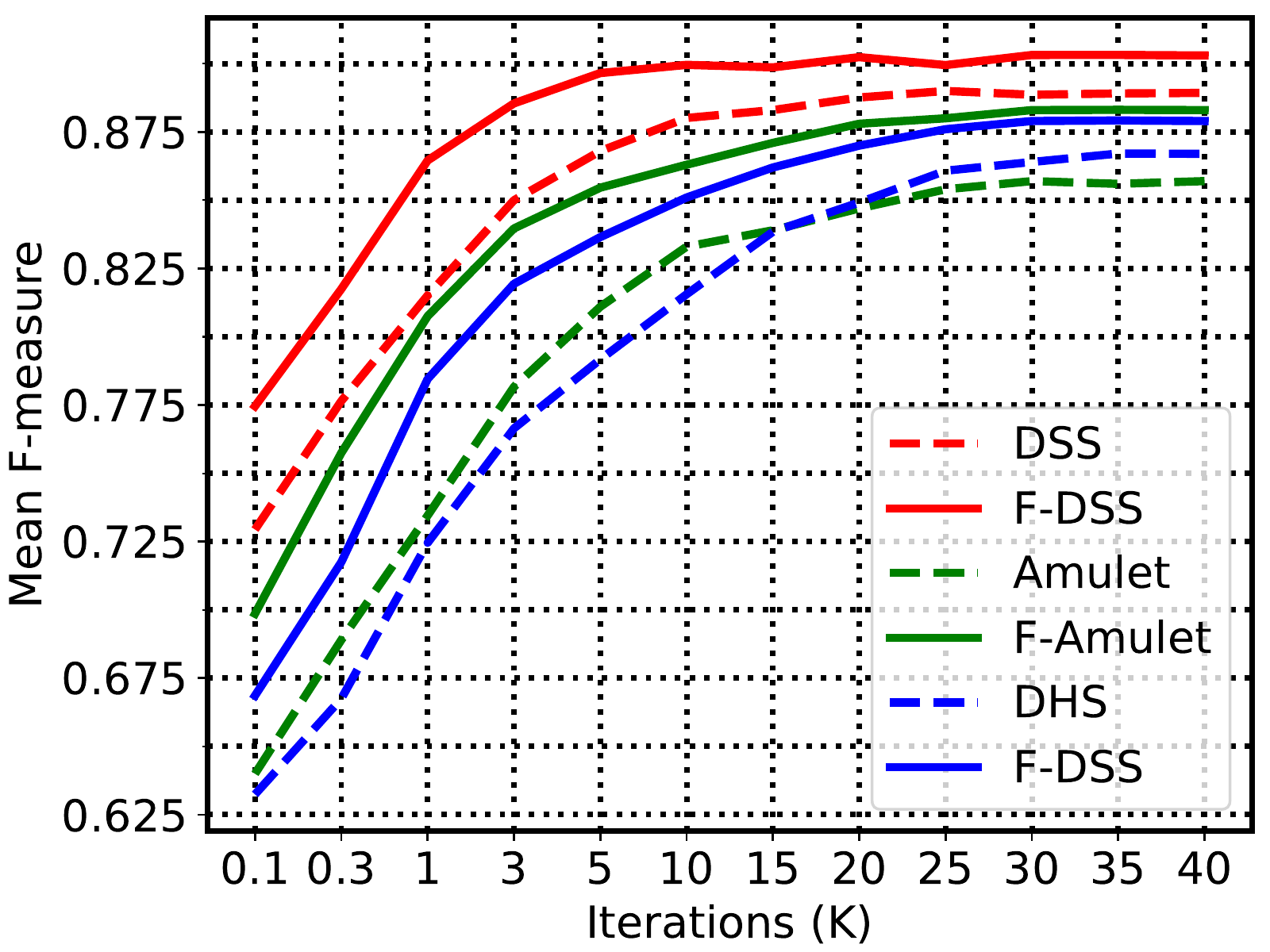}
  \end{overpic}
  \caption{Performance versus training iterations.
  Our method presents faster convergence and higher converged performance.}
  \label{fig:f-iter}
      \vspace{-10pt}
\end{figure}
}

We observe that our FLoss offers a per-iteration performance promotion for all the three saliency models.
We also find
that the FLoss-based methods quickly learn to
focus on the salient object area and achieve high F-measure score after hundreds of iterations.
While cross-entropy based methods produce blurry outputs and cannot localize
salient areas very preciously.
As shown in Fig.~\ref{fig:f-iter}, FLoss based methods converge faster than its cross entropy
competitors and get higher converged performance.

\vspace{-4pt}
\section{Conclusion}
\vspace{-4pt}
In this paper, we propose to directly maximize the F-measure for salient object detection.
We introduce the FLoss that is differentiable w.r.t the predicted posteriors
as the optimization objective of CNNs.
The proposed method achieves better performance in terms of better handling biased data distributions.
Moreover, our method is stable against the threshold and able to produce high-quality saliency maps
under a wide threshold range, showing great potential in real-world applications.
By adjusting the $\beta^2$ factor, one can easily adjust the
compromise between precision and recall,
enabling flexibility to deal with various applications.
Comprehensive benchmarks on several popular datasets illustrate the advantage of the proposed
method.

\paragraph{Future work.}
We plan to improve the performance and efficiency of the proposed method
by using recent backbone models, \eg, \cite{gao2019res2net,MobileNetV2}.
Besides, the FLoss is potentially helpful to other binary dense prediction tasks
such as edge detection~\cite{RcfEdgePami2019}, shadow detection~\cite{Hu_2018_CVPR}
and skeleton detection~\cite{zhao2018hifi}.

\paragraph{Acknowledgment.}
This research was supported by NSFC (61572264, 61620106008), 
the national youth talent support program, 
and Tianjin Natural Science Foundation (17JCJQJC43700, 18ZXZNGX00110).

{\small
\bibliographystyle{ieee_fullname}
\bibliography{fmeasure}
}

\end{document}